\documentclass[11pt,letterpaper]{article}

\usepackage[utf8]{inputenc}
\usepackage[T1]{fontenc}
\usepackage[margin=1in]{geometry}
\usepackage{graphicx}
\usepackage{float}
\usepackage{booktabs}
\usepackage{array}
\usepackage{multirow}
\usepackage{pifont}

\usepackage{makecell}
\usepackage{subcaption}
\usepackage{amsmath,amssymb}
\usepackage{tcolorbox}
\tcbuselibrary{skins,breakable}
\usepackage{enumitem}
\usepackage{soul}
\usepackage[acronym]{glossaries}
\glsdisablehyper
\usepackage{algorithm}
\usepackage[noend]{algpseudocode}
\usepackage{changepage}
\usepackage{tabularx}
\usepackage[numbers,sort&compress]{natbib}
\usepackage[hidelinks]{hyperref}
\usepackage{cleveref}

\algnewcommand{\Step}[2]{\State \textbf{Step #1:} \textsc{#2}}
\algnewcommand{\IState}[1]{\State \hspace{1.2em} #1}
\algtext*{EndProcedure}
\algtext*{EndIf}
\algtext*{EndFunction}
\algtext*{EndFor}

\newacronym{hrc}{HRC}{Human-Robot Collaboration}
\newacronym{llm}{LLM}{Large Language Model}
\newacronym{mas}{MAS}{Multi-Agent System}

\title{CoViLLM: An Adaptive Human-Robot Collaborative Assembly Framework Using Large Language Models}
\author{Jiabao Zhao \quad Jonghan Lim \quad Hongliang Li \quad Ilya Kovalenko\\
\small The Pennsylvania State University, University Park, PA, USA}
\date{}

\begin{document}
\maketitle

\begin{abstract}
With increasing demand for mass customization, traditional manufacturing robots that rely on rule-based operations lack the flexibility to accommodate customized or new product variants. Human-Robot Collaboration has demonstrated potential to improve system adaptability by leveraging human versatility and decision-making capabilities. However, existing Human-Robot Collaborative frameworks typically depend on predefined perception-manipulation pipelines, limiting their ability to autonomously generate task plans for new product assembly. In this work, we propose CoViLLM, an adaptive human-robot collaborative assembly framework that supports the assembly of customized and previously unseen products. CoViLLM combines depth-camera-based localization for object position estimation, human operator classification for identifying new components, and a Large Language Model for assembly task planning based on natural language instructions. The framework is validated on the NIST Assembly Task Board for known, customized, and new product cases. Experimental results show that the proposed framework enables flexible collaborative assembly by extending Human-Robot Collaboration beyond predefined product and task settings.
\end{abstract}


\section{Introduction}
\label{sec:intro}

The rapid development of robotic technologies has significantly improved manufacturing efficiency and productivity. 
With the increasing demand for personalized products, traditional manufacturing robots that are hard-coded are limited in their adaptability to address mass customization \cite{dhanda2025reviewing}. They often require human operators to manually update and adapt the fundamental functionality for specific customized or new products, thus reducing manufacturing efficiency. \gls{hrc} has the potential to address this challenge by integrating manufacturing robots with the versatility of human operators. The concept of \gls{hrc} refers to the interaction and cooperation between humans and robots within a shared workspace.

Previous work has focused on developing \gls{hrc} frameworks to improve productivity and efficiency while ensuring human-robot interaction safety in manufacturing assembly \cite{unhelkar2018human, guo2024fast}. In addition, recent studies have explored various \gls{hrc} approaches, including gesture-based robot instruction frameworks for collaborative assembly \cite{wang2025multi}, worker-aware mixed-reality systems for manufacturing information recommendation \cite{choi2025smart}, and scene-centric mixed-reality robot programming methods that improve segmentation in cluttered environments \cite{yin2026mixed}. However, existing \gls{hrc} systems are often complex and do not facilitate user-friendly interaction, demanding extensive specialized training in coding and robot operation~\cite{gkournelos2024llm}. Moreover, the natural language barrier when interacting with robots causes psychological stress and tension for the human operator~\cite{korner2019perceived}. Therefore, it is essential to develop a convenient human-robot communication system to facilitate a user-friendly experience and collaboration.
\begin{figure*}[!t]
    \centering
    \includegraphics[width=1\textwidth]{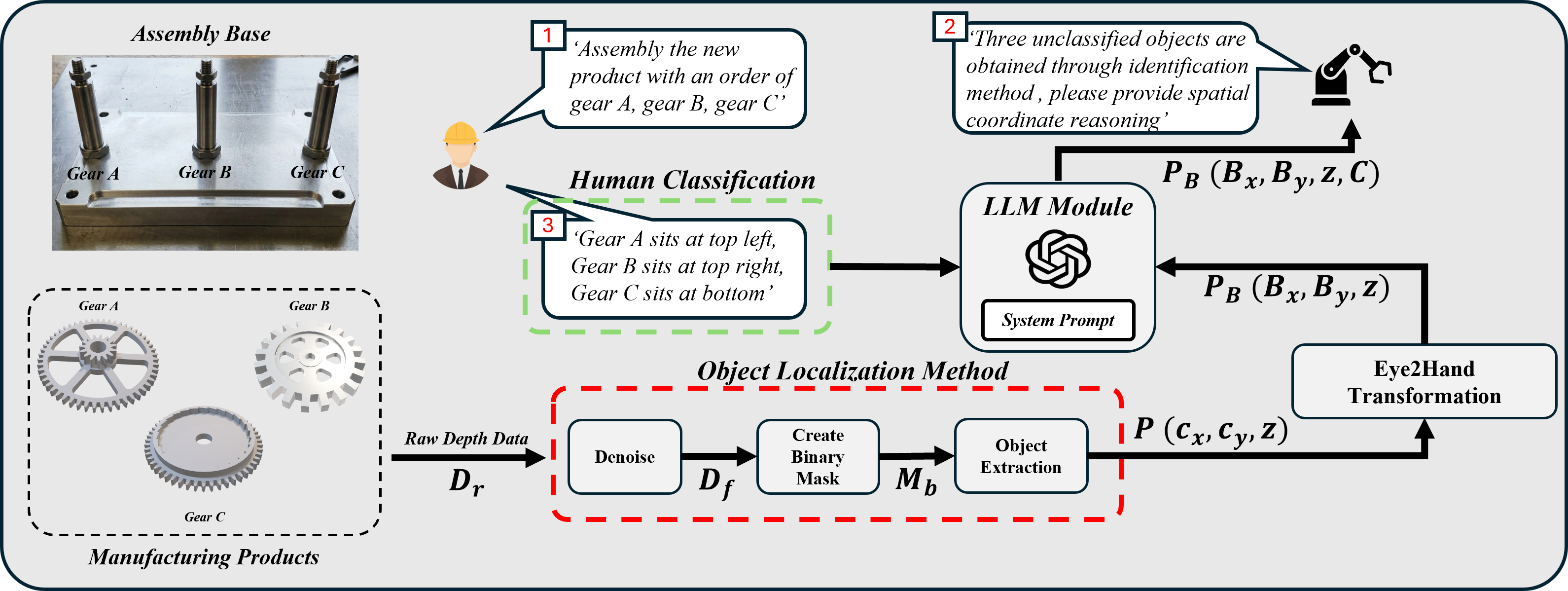}
    \caption{Overview of collaborative vision and large language model for human-robot assembly.}
    \label{fig:Overview}
\end{figure*}
Recent advances in~\glspl{llm} have sparked growing interest in their integration into~\gls{hrc}. An~\gls{llm} refers to a comprehensive system that can handle complex language tasks with reasoning, including translation, summarization, and conversational interactions \cite{naveed2025comprehensive}. \gls{llm} serves as an intermediary layer or ``translator'' between human operators and robots, enabling collaboration through natural language communication. Previous studies have demonstrated the potential of~\glspl{llm} by enabling robots to understand, process information, and make decisions based on human operator instructions \cite{mon2025embodied,singh2022progprompt,lim2024large}. Although artificial intelligence technologies such as vision-language models provide open-vocabulary recognition for new products, they remain constrained by reliability, high computational costs, and latency~\cite {zhang2024vision}. Therefore, there is a pressing need for a comprehensive \gls{hrc} framework capable of detecting customized and new products at runtime to improve the efficiency and adaptability of \gls{hrc} in manufacturing.

To address these limitations, we propose \textbf{Collaborative Vision for Human--Robot Assembly Using a Large Language Model (CoViLLM)}, as illustrated in \Cref{fig:Overview}. The proposed framework combines depth-based localization, human operator feedback for component classification, and \gls{llm}-based reasoning to enable runtime identification of previously unseen manufacturing components. Furthermore, CoViLLM converts natural-language operator instructions into executable assembly sequences. Unlike conventional \gls{hrc} systems that depend on fixed perception pipelines and predefined product knowledge, the proposed framework supports more adaptive and flexible collaborative assembly in dynamic manufacturing environments. The main contributions of this work are summarized as follows:
(1) A runtime human-in-the-loop collaborative vision framework that integrates depth-based localization, human operator classification, and \gls{llm} reasoning for the identification of previously unseen manufacturing components, (2) An \gls{llm}-enabled \gls{hrc} framework that dynamically generates assembly sequences for customized and new products from natural-language operator instructions.

The rest of this manuscript is organized as follows. Section~\ref{sec:methodology} presents the problem formulation and the proposed framework. Section~\ref{sec:case_studies} validates the proposed framework through experimental case studies using a standardized assembly task board. Finally, Section~\ref{sec:conclusion} concludes the paper and discusses future research directions.

\section{Methodology}
\label{sec:methodology}

\subsection{Problem Formulation}
Let $P = (p_1, p_2, p_3, \ldots, p_n)$ denote the ordered assembly sequence for a manufacturing product, where \(p_i\) represents the assembly subtask executed at step \(i\), for \(i=1,2,\ldots,n\), and \(n\) is the total number of subtasks required to complete the product. Each subtask \(p_i\) is characterized by two essential elements, \(L_i\) and \(C_i\), where \(L_i=(x_i,y_i,z_i)\) denotes the spatial coordinate of the target component with respect to the robot base frame, and \(C_i\) denotes the category of the component located at \(L_i\). Together, \(L_i\) and \(C_i\) enable the system to identify the correct component, execute the pick operation, and place it at the appropriate assembly location. In conventional \gls{hrc} settings, component location \({L_i}\) and component category \({C_i}\) are typically obtained using pretrained computer vision models. However, these models are limited in identifying previously unseen components that are not in the training dataset. To address this limitation, the next section presents our framework that integrates depth-based localization, human classification, and \gls{llm}-based reasoning for dynamic component identification and assembly-sequence planning.

\begin{figure*}[t!]
    \centering
    \includegraphics[width=1\textwidth]{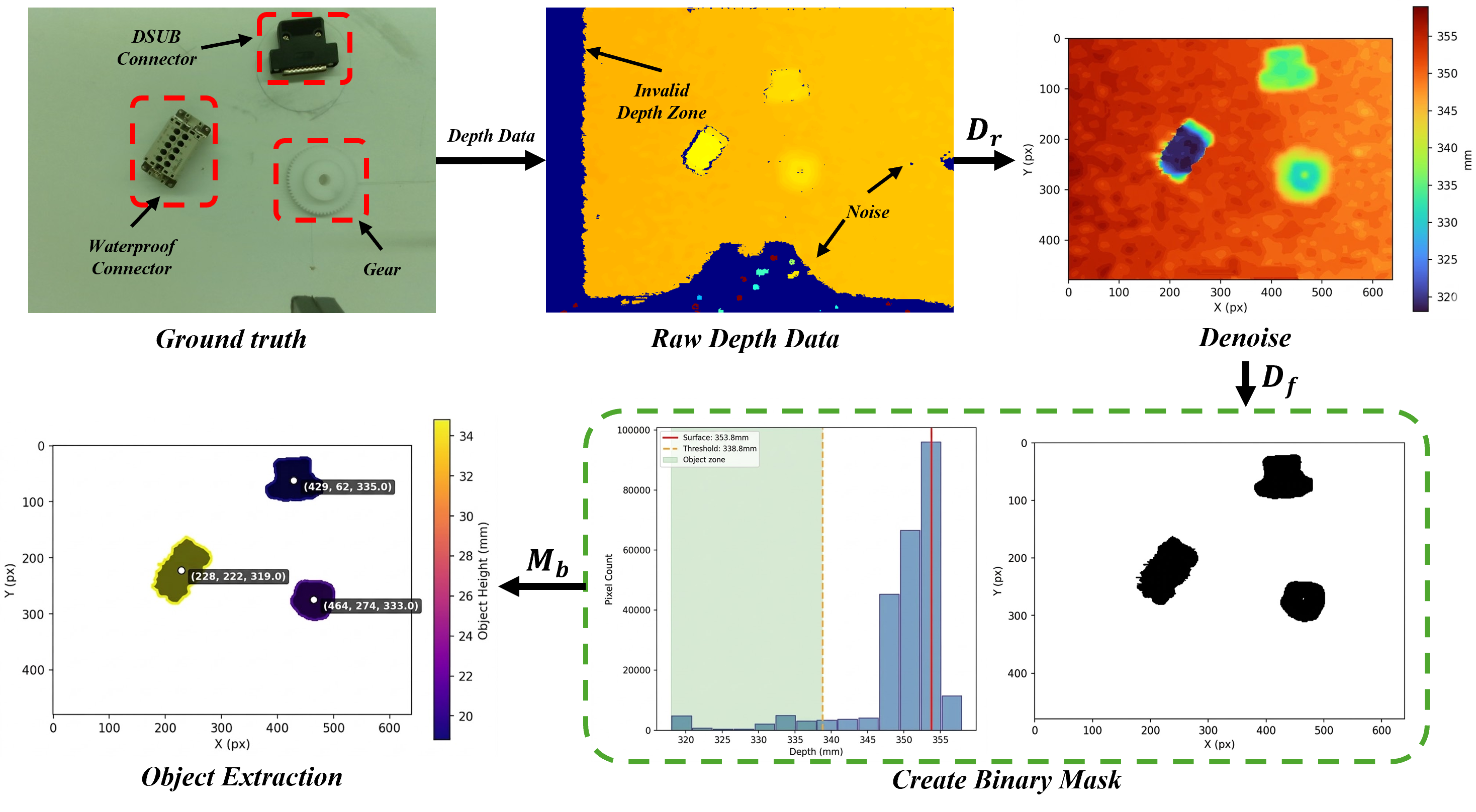}
    \caption{Object Localization through Algorithm 1.}
    \label{fig:localization}
\end{figure*}

\subsection{Localization}
The localization of unseen components is obtained through three steps: Denoise, Create Binary Mask, and Object Extraction, as shown in \Cref{alg:localization}. A raw depth map \(\boldsymbol{D}_r \in \mathbb{R}^{640 \times 480}\) is captured by the depth camera. Each element \(\boldsymbol{D}_r(x,y)\) corresponds to the depth measurement at pixel location \((x,y)\). Firstly, Denoise is applied to convert the $\boldsymbol{D}_r$ to a refined depth map $\boldsymbol{D}_f$ through a sequence of filtering operations. Specifically, spatial, temporal, and hole-filling filters are applied to improve the quality of the depth map by reducing local noise, enhancing frame stability, and decreasing invalid depth regions. To improve the effectiveness of these filtering operations, the raw depth map is first transformed into disparity space and then converted back to depth space after filtering. Secondly, Create Binary Mask step is used to segment candidate object regions $M_b$ from the refined $\boldsymbol{D}_f$. Valid depth values are extracted from a region of interest \(M\) under the constraint \(d_{\min} < d < d_{\max}\). The dominant surface depth \(d_s\) is then estimated using the histogram mode of the valid depth set. An object threshold \(t_{obj}\) is computed by subtracting the minimum object height \(h_m\) from \(d_s\). The binary mask \(M_b\) is generated by selecting pixels whose depth values are smaller than \(t_{obj}\). Lastly, Object Extraction step extracts the spatial coordinates of candidate objects from \(M_b\), \(\boldsymbol{D}_f\), and \(d_s\). The output is represented as \(P = (c_x, c_y, z)\), where \(c_x\) and \(c_y\) denote the pixel coordinates of the object centroid in the image plane, and \(z\) denotes the object depth. First, contours \(C\) are identified from the binary mask \(M_b\). For each contour \(c \in C\), the contour area \(A\) and the bounding dimensions \((w,h)\) are computed. An aspect ratio measure \(\alpha\) is then calculated and used together with the contour area to reject invalid detections. For each valid contour, the image centroid \((c_x,c_y)\) is obtained from the contour moments, and the corresponding object region \(M_{obj}\) is constructed. The object depth \(z\) is then estimated as the median depth value within \(M_{obj}\). Based on these quantities, the tuple \((c_x,c_y,z)\) is appended to the output set \(P\). \Cref{fig:localization} presents a visual example of object localization.

\begin{algorithm}[t]
\caption{Localization}
\label{alg:localization}
\begin{algorithmic}[1]
\footnotesize
\Require \texttt{pyrealsense2} library aliased as \texttt{rs}
\Step{1}{Denoise ($D_r$)}
\IState{$F_{d2z} \gets \mathrm{DisparityTransform}(T)$}
\IState{$F_{z2d} \gets \mathrm{DisparityTransform}(F)$}
\IState{$F_{spatial} \gets \mathrm{SpatialFilter}(\alpha,\delta,m)$}
\IState{$F_{temp} \gets \mathrm{TemporalFilter}()$}
\IState{$F_{hole} \gets \mathrm{HoleFillingFilter}(0)$}
\IState{$D_f \gets F_{z2d}(F_{hole}(F_{temp}(F_{spatial}(F_{d2z}(D_r)))))$}
\IState{\Return $D_f$}
\Step{2}{CreateBinaryMask ($D_f$)}
\IState{$M \gets \mathrm{ROI}(D_f)$}
\IState{$M_{valid} \gets \{d \in D_f[M] \mid d_{min} < d < d_{max}\}$}
\IState{$d_s \gets \mathrm{HistogramMode}(M_{valid})$}
\IState{$t_{obj} \gets d_s - h_m$}
\IState{$M_b \gets (D_f < t_{obj}) \land (d_{min} < D_f < d_{max})$}
\IState{\Return $M_b, d_{surf}$}
\Step{3}{ObjectExtraction ($M_b, D_f, d_s$)}
\IState{$C \gets \mathrm{FindContours}(M_b)$}
\IState{\textbf{for all} $c \in C$ \textbf{do}}
\IState{\hspace{1.2em}$A \gets \mathrm{Area}(c), \quad w,h \gets \mathrm{BoundingRect}(c)$}
\IState{\hspace{1.2em}$\alpha \gets \max(w,h)/(\min(w,h)+\epsilon)$}
\IState{\hspace{1.2em}\textbf{if} $(A_{min} \leq A \leq A_{max}) \land (\alpha \leq \alpha_{max})$ \textbf{then}}
\IState{\hspace{2.4em}$(c_x,c_y) \gets \mathrm{Moments}(c)$}
\IState{\hspace{2.4em}$M_{obj} \gets \mathrm{DrawContour}(c)$}
\IState{\hspace{2.4em}$z \gets \mathrm{Median}[D_f[M_{obj}]]$}
\IState{\hspace{2.4em}$P.\mathrm{append}([c_x,c_y,z])$}
\IState{\Return $P$}
\end{algorithmic}
\end{algorithm}

\subsection{Eye2Hand Transformation}
Eye2Hand Transformation converts the pixel coordinates of each detected component in \(P\) into its corresponding 3D coordinate in the robot base frame. The transformation can be expressed as: 
\begin{equation}
\label{eq:pixel_to_base}
{\mathbf{p}}_{B}
=
{}^{B}\mathbf{T}_{E}\,{}^{E}\mathbf{T}_{C}
\begin{bmatrix}
\frac{(c_x-p_x)}{f_x}z\\[4pt]
\frac{(c_y-p_y)}{f_y}z\\[4pt]
z\\
1
\end{bmatrix},
\end{equation}
where ${\mathbf{P}}_{B}=[B_x,\,B_y,\,z,\,1]^\top$ is the homogeneous coordinate of the component in the robot base frame. The intrinsic parameters $(f_x,f_y)$ are the focal lengths in pixel units and $(p_x,p_y)$ are the principal points. The term ${}^{E}\mathbf{T}_{C}$ denotes the fixed extrinsic transformation from the camera frame $\{C\}$ to the end-effector frame $\{E\}$: 
\begin{equation}
\label{eq:TEC}
{}^{E}\mathbf{T}_{C}=
\begin{bmatrix}
{}^{E}\mathbf{R}_{C} & {}^{E}\mathbf{t}_{C}\\
\mathbf{0}^\top & 1
\end{bmatrix},
\end{equation}
where ${}^{E}\mathbf{R}_{C}\in SO(3)$ and ${}^{E}\mathbf{t}_{C}\in\mathbb{R}^3$ are the rotation matrix and translation vector, respectively. Similarly, ${}^{B}\mathbf{T}_{E}$ denotes the time-varying transformation from the end-effector frame $\{E\}$ to the robot base frame $\{B\}$:
\begin{equation}
\label{eq:TBE}
{}^{B}\mathbf{T}_{E}=
\begin{bmatrix}
{}^{B}\mathbf{R}_{E} & {}^{B}\mathbf{t}_{E}\\
\mathbf{0}^\top & 1
\end{bmatrix},
\end{equation}
where ${}^{B}\mathbf{R}_{E}\in SO(3)$ and ${}^{B}\mathbf{t}_{E}\in\mathbb{R}^3$ are provided by the robot controller at runtime through forward kinematics. 

\subsection{Human Classification}
\begin{figure}[t!]
    \centering
    \includegraphics[width=0.46\textwidth]{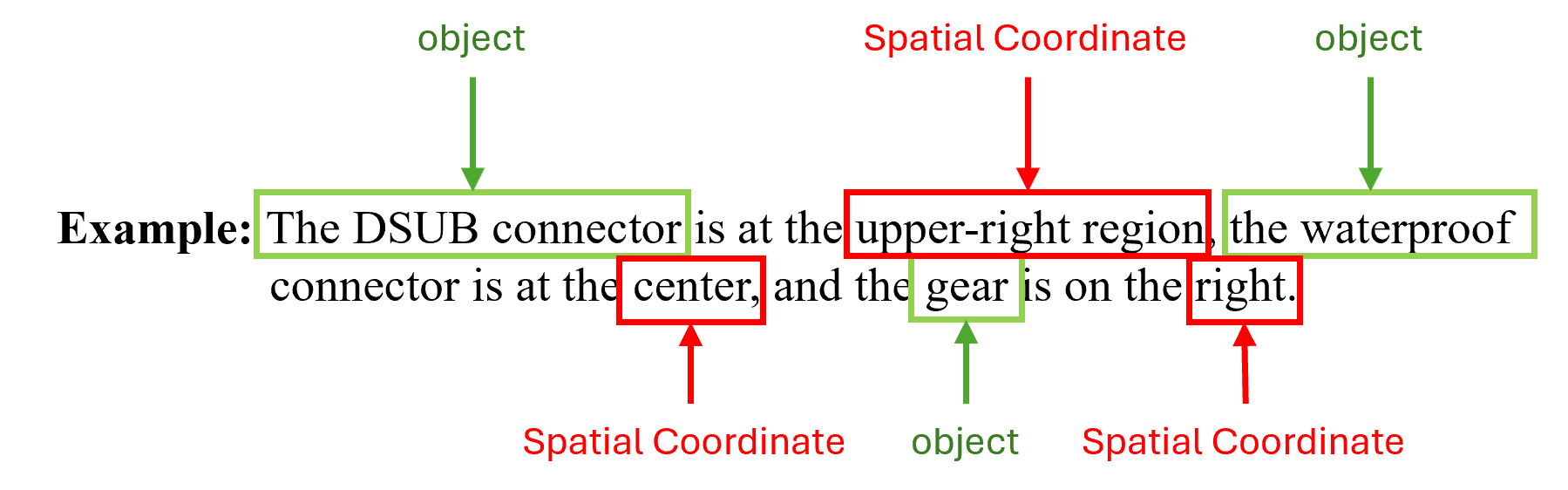}
    \caption{Human Classification Example.}
    \label{fig:fine_tuning_example}
\end{figure}

After the localization of unseen components, the human operator provides classification feedback about the spatial coordinates of the detected objects. \Cref{fig:fine_tuning_example} presents an example of this process for the localization result shown in \Cref{fig:localization}. Human classification provides cues about the relative positions of the components in the 2D workspace. Combined with localization, these classification cues are then jointly passed to the \gls{llm} module, which performs spatial reasoning to associate each semantic component description with its corresponding localized object. In this way, the human operator provides classification feedback, while the depth-based localization method extracts component coordinates, enabling collaborative runtime identification of unseen manufacturing components.

\subsection{LLM Module}
The \gls{llm} serves as the high-level reasoning module of the proposed framework. In this work, GPT-4.1 mini is used to process the human operator's request together with the component localization and human classification information and to generate a structured assembly plan through sequential reasoning. The model is configured by a system prompt that specifies the operational rules, safety constraints, and task-specific knowledge relevant to the manufacturing environment. However, generative \glspl{llm} trained on broad-domain data often exhibit limited reasoning reliability for specialized component identification tasks \cite{lim2026adaptive}. Moreover, hallucinations may reduce the robustness of deployment in practical manufacturing settings.

\begin{figure}[t!]
    \centering
    \includegraphics[width=0.46\textwidth]{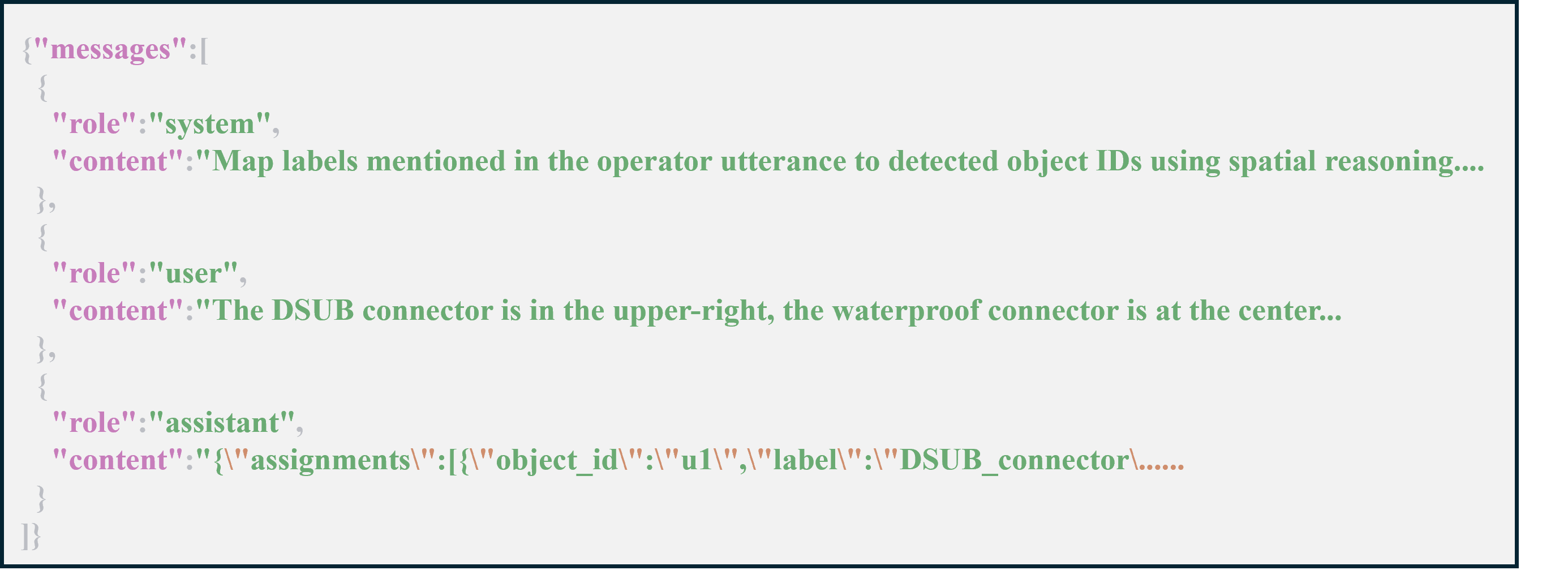}
    \caption{Fine-tuning example.}
    \label{fig:Fine-tuning_example}
\end{figure}

To improve task-specific reliability, we constructed 100 high-quality fine-tuning examples tailored to the proposed component identification framework. The dataset was designed to strengthen the model's ability to interpret localization and human classification information and to produce the corresponding structured assembly sequences. As shown in \Cref{fig:Fine-tuning_example}, each example includes a system prompt, a user input containing localization and classification information, and the corresponding assistant output. Through fine-tuning, the model learns the output structure and reasoning patterns required for this task. The training loss curve in \Cref{fig:Fine-tuning_example} shows convergence after approximately 100 steps and stabilization after about 250 steps, suggesting that the dataset provides stable learning and improved output consistency.

\begin{figure}[t!]
    \centering
    \includegraphics[width=0.46\textwidth]{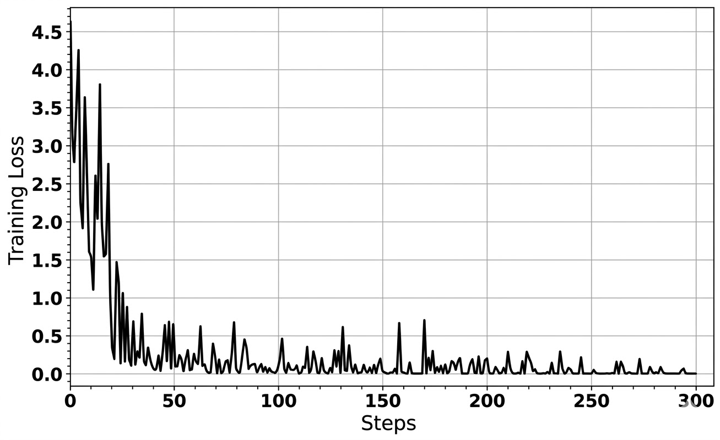}
    \caption{Training loss curve.}
    \label{fig:training_loss}
\end{figure}

\section{Case Studies}
\label{sec:case_studies}
\begin{figure}[t]
    \centering
    \includegraphics[width=0.46\textwidth]{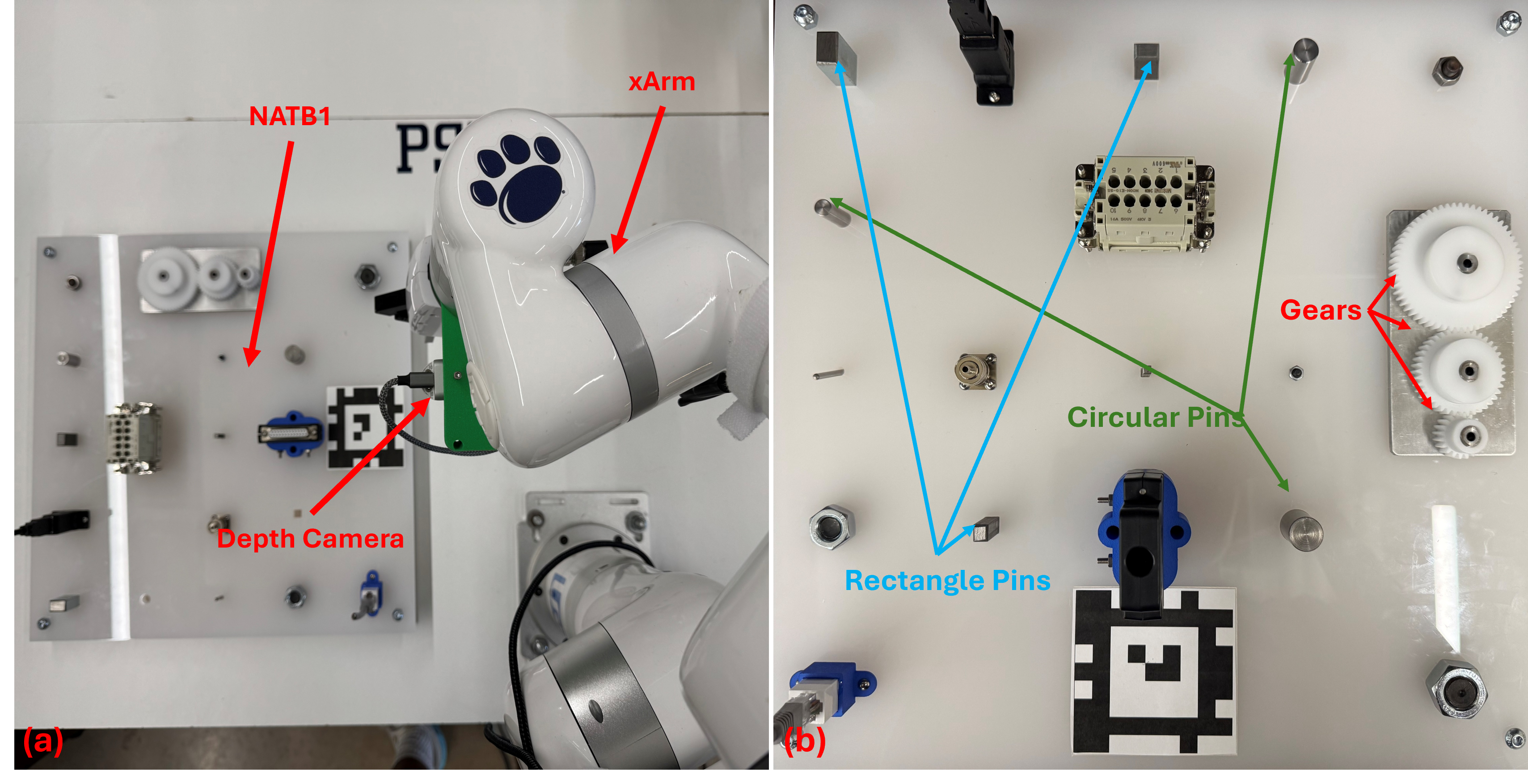}
    \caption{Overview of the experimental manufacturing setup: (a) Top-down view of the physical workspace, showing the arrangement of robot and the assembly board. (b) Detailed view of the assembly board components: gears (red arrows), circular pins (green arrows), and rectangular pins (blue arrows).}
    \label{fig:setup}
\end{figure}

\begin{table*}[t]
\centering
\caption{Three case levels and example products. The comma-separated order indicates the assembly sequence.}
\label{tab:case_levels}
\renewcommand{\arraystretch}{1.2}
\setlength{\tabcolsep}{6pt}
\begin{tabular}{@{}l p{0.27\textwidth} p{0.27\textwidth} p{0.27\textwidth}@{}}
\toprule
\textbf{Case (Level)} & \textbf{Product 1} & \textbf{Product 2} & \textbf{Product 3} \\
\midrule
\textbf{Case 1}
& small gear, small rectangular pin
& small gear, medium rectangular pin
& medium gear, small rectangular pin \\
\midrule
\textbf{Case 2}
& small gear, medium rectangular pin, medium circular pin
& medium gear, small rectangular pin, medium circular pin
& medium gear, medium circular pin, small gear \\
\midrule
\textbf{Case 3}
& big gear, small gear, small rectangular pin, small circular pin
& big circular pin, small gear, medium rectangular pin, medium circular pin
& big rectangular pin, medium gear, small rectangular pin, small gear \\
\bottomrule
\end{tabular}
\end{table*}

This section details the experimental validation of the proposed framework. We first describe the hardware testbed and the assembly task, then define the three case studies designed to evaluate the system's adaptability for different manufacturing assembly tasks.

\subsection{Experimental Setup}
The manufacturing testbed, shown in \Cref{fig:setup} (a), consists of a 6-Degree-of-Freedom (6DOF) UFactory xArm robot equipped with a UFactory xArm Gripper G2. An Intel RealSense Depth Camera D435 provides component localization via the proposed framework. The case studies are based on the NIST Assembly Task Board 1 (NATB1) \citep{nist}, a standard benchmark for robotic manipulation and assembly, as shown in \Cref{fig:setup} (b).  The NATB1 includes various components such as circular pins, rectangular pins, and gears of different sizes. In this study, components from the NATB1 are treated as new manufacturing products in the testbed. We designed three case studies with progressively increasing complexity, as shown in \Cref{tab:case_levels}, to evaluate the robustness of the proposed framework in assembling new manufacturing products. Case 1 serves as the baseline scenario and contains two unseen components. Case 2 and Case 3 contain three and four unseen components, respectively. Case 2 and 3 increase the identification and reasoning demands on the framework, evaluating its ability to assemble more complex tasks. For each case, three manufacturing products were constructed, and each product contained different unseen components and an assembly sequence.

\subsection{Results}
\begin{figure}[t]
    \centering
    \includegraphics[width=0.46\textwidth]{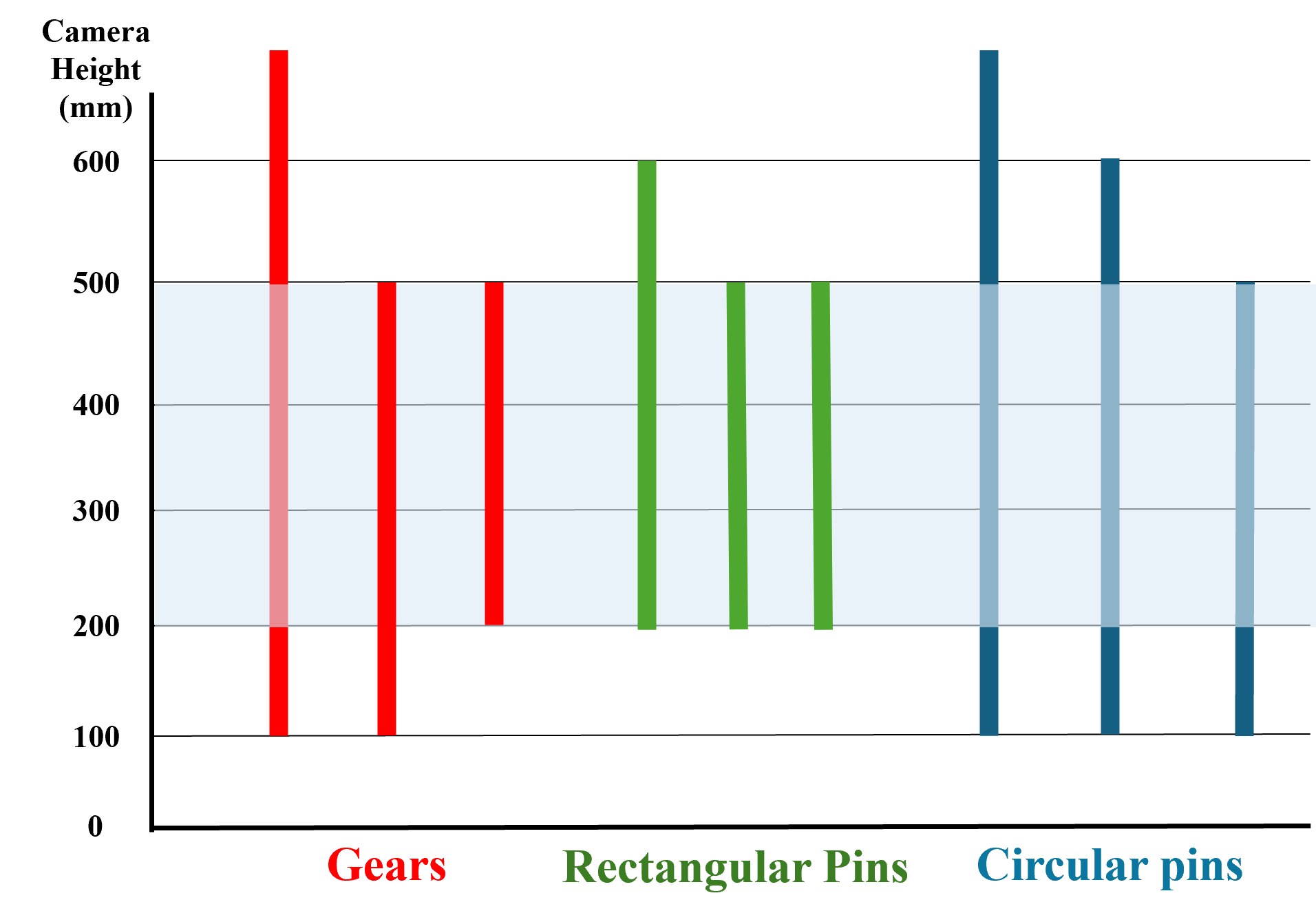}
    \caption{Valid camera-height range for effective object localization across component categories, ordered from larger to smaller component sizes within each category.}
    \label{fig:camera_height}
\end{figure}

The ability of the proposed framework to identify unseen components depends on the quality of the filtered depth frame \(D_f\), which is constrained by two physical factors. The object must have sufficient height to be distinguishable from the background in depth, and sufficient projected area to survive contour-based filtering. As \(z\) increases, both constraints become more restrictive. Therefore, selecting an appropriate camera height is essential for achieving robust localization and improving the success rate of the overall framework. Based on our experiments, the effective camera-height range is shown in \Cref{fig:camera_height}. In this paper, we set the camera height to 400 mm above the background surface across the case studies. 

\newcolumntype{Y}{>{\centering\arraybackslash}X}
\newcolumntype{L}{>{\raggedright\arraybackslash}X}

\begin{table}[t]
\centering
\caption{Task planning accuracy across the base and fine-tuned models. Each entry denotes the number of correctly generated assembly sequences out of three trials.}
\label{tab:llm_eval}
\small
\setlength{\tabcolsep}{5pt}
\renewcommand{\arraystretch}{1.15}
\begin{tabularx}{\linewidth}{@{}L Y Y Y@{}}
\toprule
\textbf{Case} & \textbf{GPT-4.1 mini} & \textbf{GPT-4.1} & \textbf{GPT-4.1 mini (FT)} \\
\midrule
Case 1 & 3/3 & 3/3 & \textbf{3/3} \\
Case 2 & 2/3 & 3/3 & \textbf{3/3} \\
Case 3 & 1/3 & 2/3 & \textbf{3/3} \\
\bottomrule
\end{tabularx}
\end{table}

To examine the robustness of the framework, we compared the performance of the fine-tuned GPT-4.1 mini model against the baseline GPT-4.1 mini and GPT-4.1 models. Table~\ref{tab:llm_eval} shows the accuracy of the \gls{llm} module in generating correct assembly task plans with the structured set $P_B$. The baseline GPT-4.1 mini and GPT-4.1 models show degraded performance when four unseen components are present in the workspace. In contrast, the fine-tuned model achieves 100\% accuracy in generating the structured output. In these cases, the baseline models often hallucinate and mismatch localization results with human classification when more unseen manufacturing products are introduced. Although all models produced correct task plans, the baseline models struggled to correctly associate localization with classification for unseen products.

Human classification information is essential for identifying unseen components. To correctly associate an unseen component with its localized position, the classification provided by the human operator must be spatially consistent with the localization results. In addition to lower accuracy, the baseline models also exhibit higher latency as the number of unseen components in the workspace increases. For each unseen component, the model must map the human classification to the corresponding pixel coordinates and reason over the relative spatial relationships among the localized objects. As a result, the baseline models require more complex spatial reasoning rather than simply following a learned output pattern, which leads to increased response latency. In comparison, the fine-tuned model exhibits lower latency because it has learned the task-specific reasoning pattern from the fine-tuning dataset.

\section{Conclusion}
\label{sec:conclusion}
The increasing demand for mass customization presents a significant challenge for existing manufacturing systems, in which pre-programmed robots struggle to accommodate personalized product requirements. In this work, we proposed \textbf{CoViLLM}, an adaptive \gls{hrc} assembly framework designed to address this gap and enhance the flexibility of manufacturing systems. The proposed framework employs an \gls{llm}-enabled agent for dynamic sequence and subtask planning, reducing reliance on predefined assembly instructions. In addition, we developed a collaborative identification system that couples localization from a depth camera with human operator classification at runtime. This integration improves the flexibility and adaptability of traditional \gls{hrc} systems. Our system successfully assembles new products with components and assembly sequences that were not predefined in its original knowledge base.

Future work will focus on equipping the proposed framework with memory capabilities to store unseen product identifications, further improving manufacturing efficiency. In addition, we will extend and evaluate the framework in scenarios where the human operator cannot provide concise classification information. Collectively, this future work will enhance the deployment of the proposed framework in real manufacturing settings.

\section*{Acknowledgments}
This research was supported by the Manufacturing PA Innovation Program.
The authors would also like to thank Dana Smith and Brent Albert (DMI Companies, Inc.) for valuable feedback regarding the proposed framework and case study.

\bibliographystyle{unsrtnat}
\bibliography{Jiabao_Zhao}

\end{document}